\documentclass[10pt,twocolumn,letterpaper]{article}

\usepackage{multirow}
\usepackage{wacv}              

\definecolor{wacvblue}{rgb}{0.21,0.49,0.74}
\usepackage[pagebackref,breaklinks,colorlinks,allcolors=wacvblue]{hyperref}

\def\wacvPaperID{793} 
\def\confName{WACV}
\def\confYear{2027}

\title{CPVB-RG: Cross-Prior Visual Bridging for \\Prior-Aware Chest X-Ray Report Generation}

\author{
Sike Xiang$^{1}$ \quad
Shuang Chen$^{1}$ \quad
Julie Cox$^{2}$ \quad
Amir Atapour-Abarghouei$^{1}$\thanks{Corresponding author.}\\
$^{1}$Department of Computer Science, Durham University, UK\\
$^{2}$Department of Radiology, County Durham and Darlington NHS Foundation Trust, UK\\
{\tt\small \{sike.xiang, shuang.chen, amir.atapour-abarghouei\}@durham.ac.uk}\\
{\tt\small j.cox1@nhs.net}
}

\begin{document}
\maketitle

\begin{abstract}
Radiologists interpret follow-up chest X-rays by comparing current and prior examinations to identify new, resolved or progressing findings. Most report generation systems either omit this longitudinal context or fuse prior features without distinguishing interval changes from anatomy that remains stable over time. We propose CPVB-RG, a prior-aware framework that combines current and prior radiographs with available clinical text. Its Cross-Prior Visual Bridge (CPVB) contains Progression-Guided Focus Attention (PGFA) and two Cross-Attention units. PGFA is designed to contrast progression-difference attention with shared-spatial attention, aiming to reduce invariant responses whilst retaining coarse spatial correspondence between examinations. The Cross-Attention units then connect the resulting longitudinal visual evidence with clinical semantics. CPVB-RG is trained in two stages, using prior-guided image-text alignment followed by autoregressive report generation with a lightweight language decoder. Experiments on MIMIC-CXR and IU X-Ray show consistent improvements across language and clinical metrics. On MIMIC-CXR, CPVB-RG achieves 0.2847 F1RadGraph and 0.6692 VLScore, with statistically significant gains over the closest decoder-matched prior-aware baseline. These results support the effectiveness of the proposed difference-aware and spatially guided fusion design for prior-aware chest X-ray report generation.
\end{abstract}

\section{Introduction}
\label{sec:intro}

Chest X-rays are among the most commonly used imaging examinations for assessing thoracic disease~\cite{ChestXray8, mimiccxr}. Automatic report generation could reduce repetitive reporting effort, but a follow-up report is not simply a textual description of the current image. Findings are often subtle and spatially localised, and their interpretation depends on whether they are new, stable or changing relative to an earlier examination~\cite{AIindI, SONG202561}. Report-generation models must therefore identify interval changes while filtering responses from anatomy that remains unchanged.

Vision-language methods learn joint representations of radiographs and clinical text for report generation~\cite{tiu2022expert, zhang2023pmcvqa, wang2024interactive}. Most generate reports from the current examination alone and cannot directly assess temporal change. Prior-aware methods such as PriorRG~\cite{liu2026priorrgpriorguidedcontrastivepretraining} incorporate earlier studies, but their current and prior features are typically combined through feature fusion or attention-based interaction. Progression-related differences and spatial responses shared across examinations are not explicitly separated. This leaves stable anatomical patterns within the fused representation, where they may obscure small but clinically relevant changes~\cite{tian2023surveydeeplearningmultimodal}.

CPVB-RG addresses this limitation by combining current radiographs, prior radiographs and available clinical text. Its central component, the Cross-Prior Visual Bridge (CPVB), contains Progression-Guided Focus Attention (PGFA) followed by two Cross-Attention units. PGFA uses separate progression-difference and shared-spatial attention branches, with the aim of reducing invariant responses whilst retaining coarse spatial correspondence between examinations. The Cross-Attention units then associate the resulting longitudinal visual evidence with clinical semantics. The model is trained through prior-guided image-text alignment before generating reports autoregressively with a lightweight language decoder.

We evaluate CPVB-RG on MIMIC-CXR~\cite{mimiccxr}, IU X-Ray~\cite{IUxray} and 135 clinical cases. On MIMIC-CXR, it achieves 0.176 BLEU-4 and 0.515 clinical efficacy F1. Under the ReXrank MIMIC-CXR Findings protocol~\cite{rexrank}, CPVB-RG obtains 0.2847 F1RadGraph~\cite{f1radgraph} and 0.6692 VLScore~\cite{vlscore}. These results exceed the closest decoder-matched prior-aware baseline by 0.0090 and 0.0582, respectively, with both differences being statistically significant. Component-level ablations examine CPVB, PGFA and their attention operations, while the clinical case analysis provides preliminary evidence beyond public benchmarks.

Our contributions are:
\begin{enumerate}[label=(\roman*)]

\item We introduce CPVB, whose PGFA module is designed to distinguish progression-related differences from shared spatial responses. This design aims to reduce invariant responses whilst retaining coarse correspondence between current and prior examinations (Figs.~\ref{CPVB} and~\ref{pgfa}).

\item We integrate CPVB into a two-stage report-generation framework. Prior-guided image-text alignment learns longitudinal multimodal representations that condition a lightweight 82M-parameter decoder (Sec.~\ref{sec:method}).

\item We evaluate CPVB-RG on two public benchmarks and 135 clinical cases, with component-level ablations. It achieves the best NLG results among the compared methods and statistically significant gains over the closest decoder-matched prior-aware baseline on clinically grounded metrics (Secs.~\ref{sec:exp:quantitative} and~\ref{sec:exp:cca}).

\end{enumerate}

\section{Related Work}
\label{sec:relatedwork}

\subsection{Chest X-Ray Vision-Language Learning and Report Generation}
\label{sec:related:medical}

\begin{figure*}[t]
  \centering
  \includegraphics[width=0.7\linewidth]{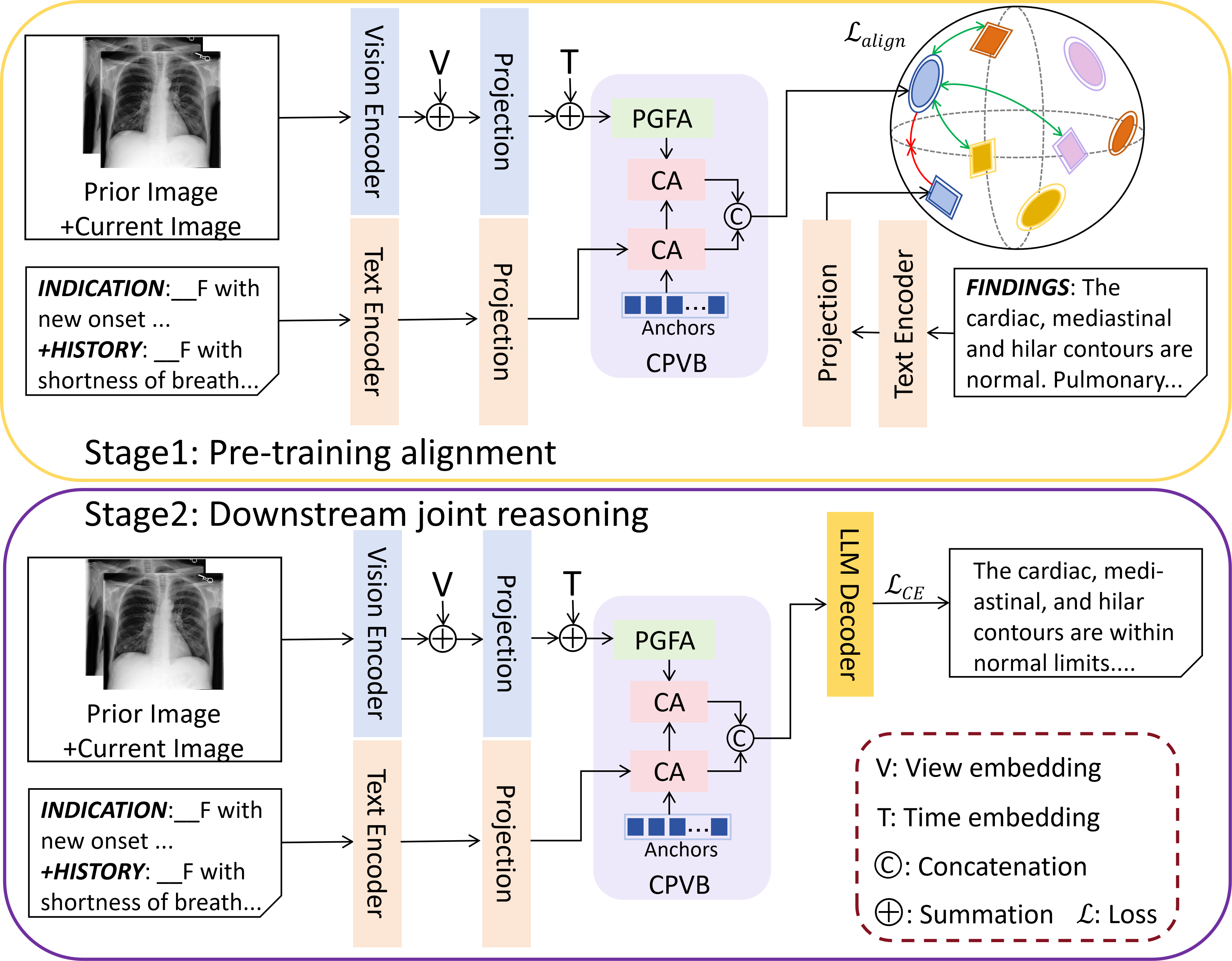}
  \caption{Overview of CPVB-RG. Current and prior chest X-rays and clinical text are first encoded, after which the Cross-Prior Visual Bridge fuses their representations for prior-guided image-text alignment and autoregressive report generation. CPVB denotes the Cross-Prior Visual Bridge (Fig.~\ref{CPVB}), PGFA denotes Progression-Guided Focus Attention (Fig.~\ref{pgfa}), and CA denotes Cross-Attention.}
  \label{overallxray}
\end{figure*}

Chest radiographs often contain low-contrast findings and ambiguous visual cues, making visual representation quality important for downstream multimodal tasks~\cite{electronics13234688}. Early systems commonly used convolutional networks, including DenseNet-based models such as CheXNet~\cite{8099726, rajpurkar2017chexnetradiologistlevelpneumoniadetection}. Vision Transformers later improved global structural modelling through token-based attention~\cite{vit, takahashi2024vitcnn}, while contrastive frameworks such as CLIP aligned visual and textual representations in a shared embedding space~\cite{clip}. General vision-language architectures, including ALBEF~\cite{ALBEF} and ViLT~\cite{vilt}, model multimodal interactions, whereas BLIP-2~\cite{blip2} and LLaVA~\cite{llava1.5} connect projected visual tokens to generative language decoders. Domain-specific language models such as Med-PaLM~\cite{medpalm}, PMC-LLaMA~\cite{pmcllama} and BioGPT~\cite{biogpt} further support medical terminology and clinical language modelling. Recent chest X-ray systems, including CheXagent~\cite{chexagent}, MAIRA-2~\cite{maira2} and UniRG-CXR~\cite{unirg}, use larger multimodal decoders and broader training objectives. These systems strengthen image-text reasoning and report generation but do not explicitly separate temporal changes from anatomical responses shared across examinations.

\subsection{Prior-Aware and Longitudinal Chest X-Ray Modelling}
\label{sec:related:longitudinal}

Longitudinal report-generation methods differ in how they use information from earlier patient visits. Zhu et al.~\cite{zhu2023prefill} combine current and prior chest X-rays with the previous report to pre-fill the Findings section. HERGen~\cite{wang2024hergen} integrates multi-visit records using a group causal transformer and curriculum learning, while HC-LLM~\cite{liu2025hcllm} uses time-shared and time-specific representations with multimodal constraints to guide an LLM. MLRG~\cite{liu2025mlrg} uses multi-view longitudinal contrastive learning, while BiOTPrompt~\cite{liu2026biotprompt}, TIM~\cite{dong2026tim} and MARE~\cite{gao2026mare} explore transport-based, decoupled and analogical temporal reasoning. PriorRG~\cite{liu2026priorrgpriorguidedcontrastivepretraining}, the closest decoder-matched prior-aware method, combines prior-guided contrastive pre-training, coarse-to-fine decoding and attention-based temporal fusion. TILA~\cite{tila} learns directional changes for progression classification and retrieval, while TRACE~\cite{trace} generates spatially grounded change descriptions. In contrast, CPVB-RG models current--prior interactions through separate progression-difference and shared-spatial attention branches and integrates the resulting evidence with clinical text for full-report generation.

\section{Method}
\label{sec:method}

CPVB-RG is a two-stage framework for prior-aware chest X-ray report generation, as shown in Fig.~\ref{overallxray}. Its inputs are a current examination, an available prior examination and clinical context such as indication or history. The visual inputs are encoded with a shared vision encoder, while the clinical context is processed by a text encoder. Their projected representations are fused by the Cross-Prior Visual Bridge (CPVB). During Stage 1, the fused representation is aligned with the corresponding Findings text, which is used only as contrastive supervision and is unavailable at inference. During Stage 2, the aligned representation conditions a lightweight language decoder for autoregressive report generation. This separation prevents the reference Findings text from entering the inference pipeline.

\subsection{Prior-Guided Image-Text Alignment Pre-training}
\label{sec:method:PGIA}

The first stage learns longitudinal multimodal representations from the current examination, the constructed prior and the clinical context. The current and prior chest X-rays are encoded by a shared vision encoder, after which view-position embeddings are added to retain acquisition-specific information. The visual features are projected to a common feature dimension, and temporal embeddings distinguish the current examination from its prior. The clinical context is independently encoded and projected to the same dimension before multimodal interaction.

For each current examination, a valid prior is a strictly earlier chest X-ray examination from the same patient. If multiple earlier examinations are available, we select the chronologically closest one without imposing a minimum time interval. A matching view is prioritised when available. Otherwise, the available view is retained and identified by its view-position embedding. All examinations undergo the same preprocessing, without additional subjective exclusion based on image quality. When no valid prior exists, the current visual tokens, after adding the current temporal embedding, are duplicated as the prior placeholder without introducing a prior temporal embedding or information from another patient. Although the branch-specific projections may produce nonzero differences, both branches depend solely on the current examination. PGFA therefore acts as a learned current-only refinement rather than providing evidence of longitudinal change. IU X-Ray is trained and evaluated consistently under this setting.

\begin{figure}[t]
  \centering
  \large
  \includegraphics[width=0.85\linewidth]{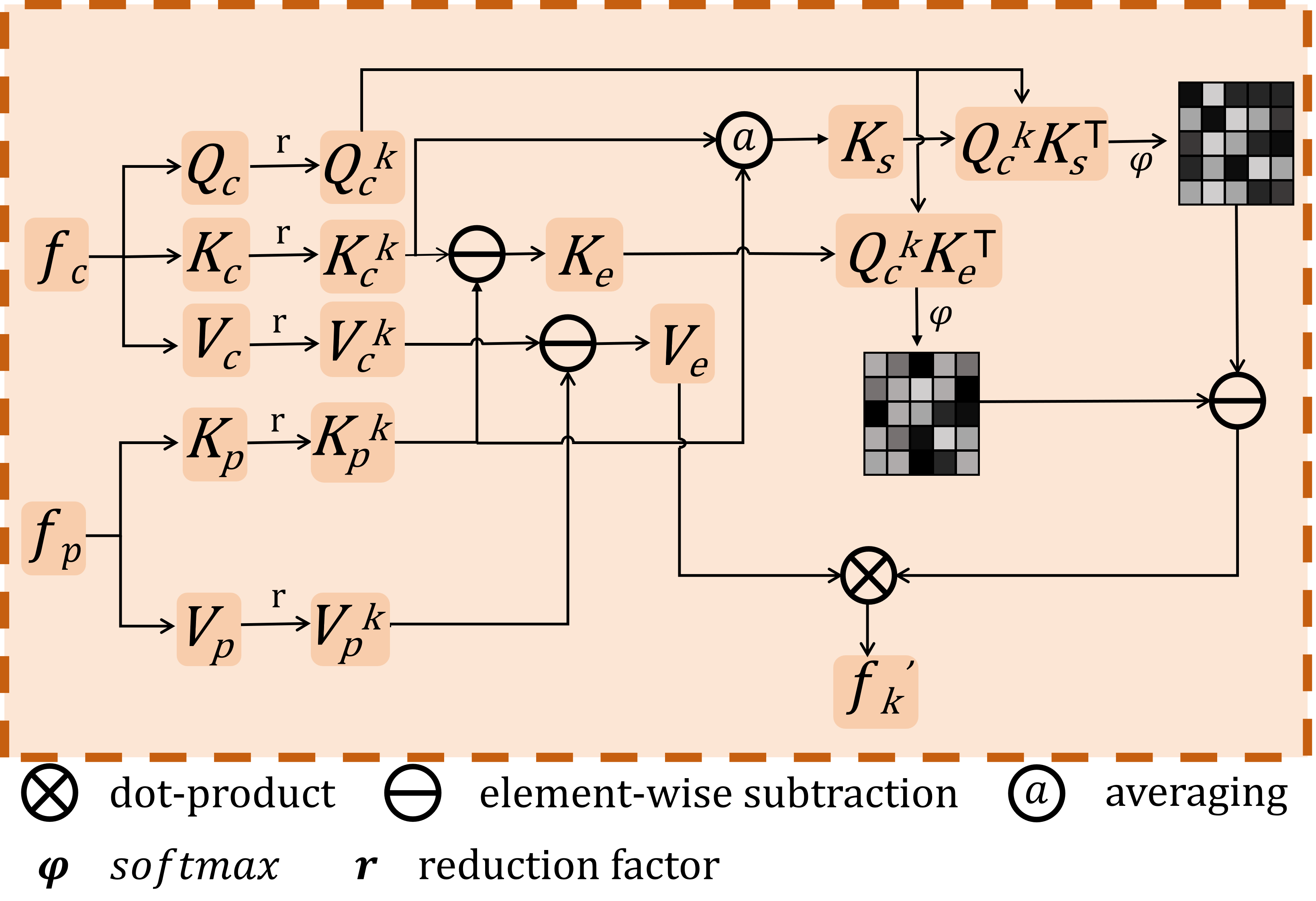}
  \caption{Structure of the PGFA module.}
  \label{pgfa}
\end{figure}

CPVB connects longitudinal visual evidence with clinical context. It contains a Progression-Guided Focus Attention (PGFA) module and two Cross-Attention (CA) units, as shown in Figs.~\ref{CPVB} and~\ref{pgfa}. PGFA first constructs separate branches to model progression-related differences and spatial responses shared by the current and prior examinations. The CA units subsequently associate these visual features with the encoded clinical context.

Let $\mathbf{f}_c,\mathbf{f}_p\in\mathbb{R}^{B\times N\times C}$ denote the current and prior visual tokens, where $B$, $N$ and $C$ are the batch size, token count and feature dimension. To reduce computation, average pooling with reduction factor $r$ is applied along the token dimension:

{\small
\begin{equation}
\bar{\mathbf{f}}_c=\mathcal{R}_r(\mathbf{f}_c),\qquad
\bar{\mathbf{f}}_p=\mathcal{R}_r(\mathbf{f}_p),
\end{equation}
}

where $\bar{\mathbf{f}}_c,\bar{\mathbf{f}}_p\in
\mathbb{R}^{B\times N_r\times C}$ and
$N_r=\lceil N/r\rceil$. PGFA then computes the current query, key and value representations and the prior key and value representations:

{\small
\begin{equation}
\mathbf{Q}_c=\bar{\mathbf{f}}_c\mathbf{W}_Q,\qquad
\mathbf{K}_c=\bar{\mathbf{f}}_c\mathbf{W}_{K_c},\qquad
\mathbf{V}_c=\bar{\mathbf{f}}_c\mathbf{W}_{V_c},
\end{equation}
}

{\small
\begin{equation}
\mathbf{K}_p=\bar{\mathbf{f}}_p\mathbf{W}_{K_p},\qquad
\mathbf{V}_p=\bar{\mathbf{f}}_p\mathbf{W}_{V_p},
\end{equation}
}

where the projection matrices are learned. The projected features are divided into $h$ attention heads with head dimension $d_h=C/h$. The progression-difference branch subtracts the prior keys and values from their current counterparts:

{\small
\begin{equation}
\mathbf{K}_e=\mathbf{K}_c-\mathbf{K}_p,\qquad
\mathbf{V}_e=\mathbf{V}_c-\mathbf{V}_p.
\end{equation}
}

The shared-spatial branch averages the current and prior keys:

{\small
\begin{equation}
\mathbf{K}_s
=
\frac{\mathbf{K}_c+\mathbf{K}_p}{2}.
\end{equation}
}

This averaging uses the approximately consistent thoracic layout across examinations and is intended to form a symmetric representation of shared anatomical responses whilst reducing examination-specific noise. Querying $\mathbf{K}_s$ is equivalent to averaging the current-key and prior-key similarity logits before softmax. The resulting attention therefore provides coarse token-level spatial guidance without assuming strict pixel-level registration. The progression-difference and shared-spatial attention maps are computed as:

\begin{figure}[t]
  \centering
  \large
  \includegraphics[width=0.85\linewidth]{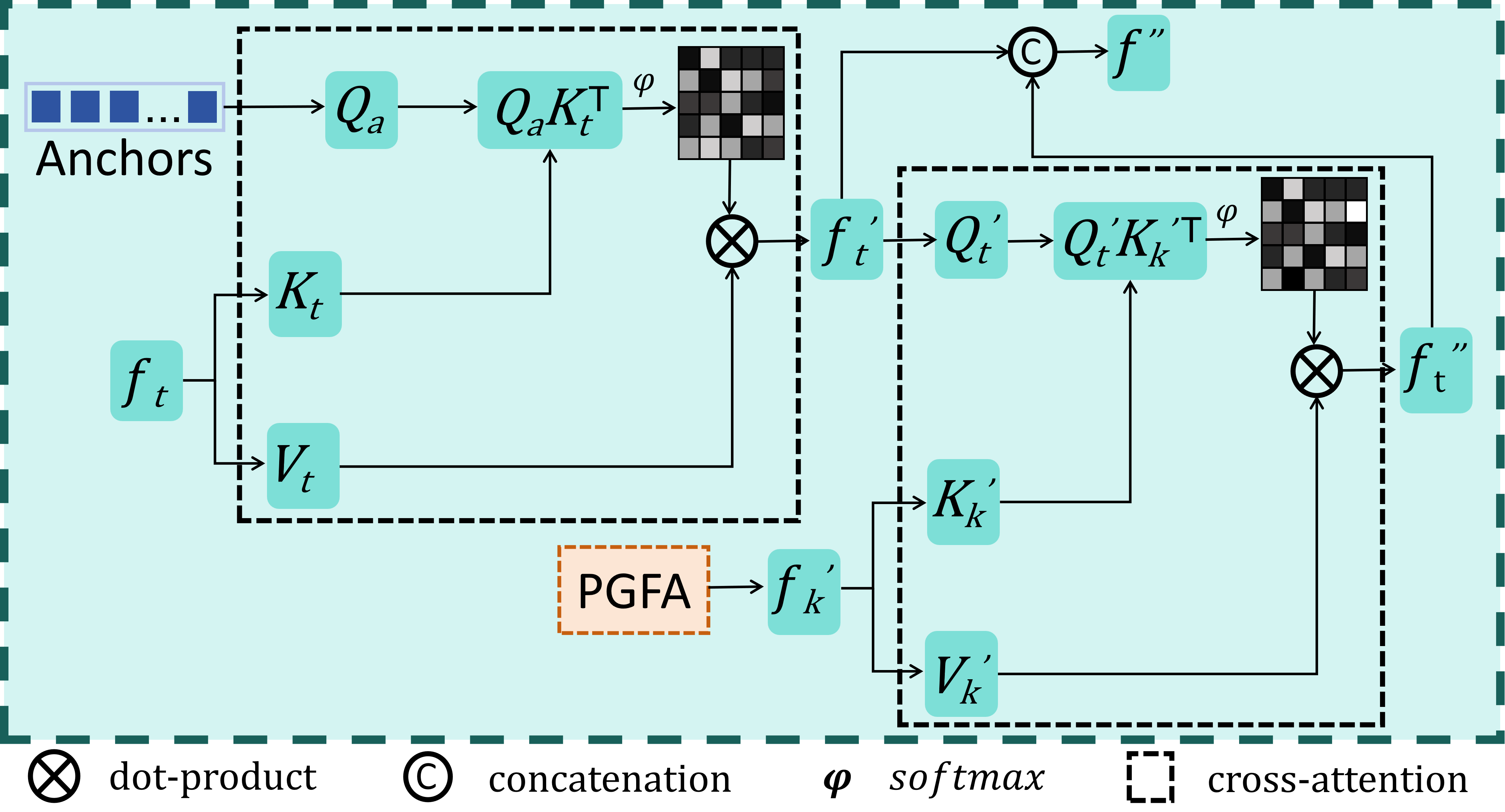}
  \caption{Structure of the CPVB module.}
  \label{CPVB}
\end{figure}

{\small
\begin{equation}
\begin{aligned}
\mathbf{A}_e
&=
\operatorname{softmax}
\left(
\frac{\mathbf{Q}_c\mathbf{K}_e^{\top}}{\sqrt{d_h}}
\right),\\
\mathbf{A}_s
&=
\operatorname{softmax}
\left(
\frac{\mathbf{Q}_c\mathbf{K}_s^{\top}}{\sqrt{d_h}}
\right).
\end{aligned}
\end{equation}
}

Their difference is applied to the progression values:

{\small
\begin{equation}
\tilde{\mathbf{f}}_c
=
\left(
\mathbf{A}_e-\lambda\mathbf{A}_s
\right)\mathbf{V}_e,
\end{equation}
}

where $\lambda=\operatorname{sigmoid}(\alpha)$ is a learned balancing coefficient. After head merging and output projection, the reduced tokens are interpolated to the original token count. A residual connection and layer normalisation produce the PGFA output:

{\small
\begin{equation}
\mathbf{f}'_c
=
\operatorname{LN}
\left(
\mathbf{f}_c+
\mathcal{U}_r
\left(
\tilde{\mathbf{f}}_c\mathbf{W}_O
\right)
\right),
\end{equation}
}

where $\mathcal{U}_r$ denotes token interpolation and $\mathbf{W}_O$ is the output projection. Three PGFA layers are stacked to refine the current visual tokens, while the prior tokens remain fixed as longitudinal guidance.

The two CA units use learnable anchor tokens to connect the PGFA output with clinical context. The first CA unit uses the anchors as queries over the clinical tokens, producing a text-conditioned anchor representation $\mathbf{F}_t$. The second uses $\mathbf{F}_t$ to retrieve longitudinal visual evidence from $\mathbf{f}'_c$, producing $\mathbf{F}_v$. Their concatenation forms the CPVB representation:

{\small
\begin{equation}
\mathbf{F}_b
=
\operatorname{LN}
\left(
[\mathbf{F}_t;\mathbf{F}_v]
\right).
\end{equation}
}

During Stage 1, the CPVB representation is aligned with the Findings text. Let $T_f$ denote the Findings section used as contrastive supervision. The visual and textual embeddings are:

{\small
\begin{equation}
\mathbf{z}_v
=
\operatorname{Norm}
\left(
\operatorname{Mean}(\mathbf{F}_b)
\right),
\qquad
\mathbf{z}_t
=
\operatorname{Norm}
\left(
E_t(T_f)_{\mathrm{CLS}}
\right).
\end{equation}
}

A patient-aware InfoNCE objective aligns the two embedding spaces, treating samples from the same patient as positives. Specifically, for each visual sample, all Findings texts from the same patient within the minibatch are treated as positives, including the matched report of the current examination and reports from other visits when present. Texts from different patients are treated as negatives, and the target weights are normalised across all positives. The resulting objective is:

{\small
\begin{equation}
\mathcal{L}_{\mathrm{align}}
=
\mathcal{L}_{\mathrm{InfoNCE}}
\left(
\mathbf{z}_v,\mathbf{z}_t
\right).
\end{equation}
}

Findings text is used only for this objective and is not supplied to CPVB or the report decoder during inference.

\begin{table*}[t]
\centering
\small

\begin{tabular}{l|l|l|cccccc}
\toprule
\textbf{Dataset} & \textbf{Methods} & \textbf{Year} & \textbf{BLEU-1} & \textbf{BLEU-2} & \textbf{BLEU-3} & \textbf{BLEU-4} & \textbf{ROUGE-L} & \textbf{METEOR} \\
\midrule
\multirow{19}{*}{\parbox[c]{1.2cm}{\centering MIMIC-CXR}}
& R2Gen~\cite{r2gen} & 2020 & 0.353 & 0.218 & 0.145 & 0.103 & 0.277 & 0.142 \\
& R2GenCMN~\cite{r2gencmn} & 2021 & 0.353 & 0.218 & 0.148 & 0.106 & 0.278 & 0.142 \\
& PPKED~\cite{ppked} & 2021 & 0.360 & 0.224 & 0.149 & 0.106 & 0.284 & 0.149 \\
& AlignTrans~\cite{aligntrans} & 2021 & 0.378 & 0.235 & 0.156 & 0.112 & 0.283 & 0.158 \\
& CMCL~\cite{cmcl} & 2021 & 0.344 & 0.217 & 0.140 & 0.097 & 0.281 & 0.133 \\
& Clinical-BERT~\cite{clinicalbert} & 2022 & 0.383 & 0.230 & 0.151 & 0.106 & 0.275 & 0.144 \\
& METformer~\cite{metformer} & 2023 & 0.386 & 0.250 & 0.169 & 0.124 & 0.291 & 0.152 \\
& DCL~\cite{dcl} & 2023 & -- & -- & -- & 0.109 & 0.284 & 0.150 \\
& R2GenGPT~\cite{r2gengpt} & 2023 & 0.408 & 0.256 & 0.174 & 0.125 & 0.285 & 0.167 \\
& KiUT~\cite{kiut} & 2023 & 0.393 & 0.243 & 0.159 & 0.113 & 0.285 & 0.160 \\
& BootstrappingLLM~\cite{bootllm} & 2024 & 0.402 & 0.262 & 0.180 & 0.128 & 0.291 & 0.175 \\
& CoFE~\cite{cofe} & 2024 & -- & -- & -- & 0.125 & 0.304 & 0.176 \\
& MAN~\cite{man} & 2024 & 0.396 & 0.244 & 0.162 & 0.115 & 0.274 & 0.151 \\
& DCG~\cite{DCG} & 2024 & 0.397 & 0.258 & 0.166 & 0.126 & 0.295 & 0.162 \\
& Med-LMM~\cite{medlmm} & 2024 & -- & -- & -- & 0.128 & 0.289 & 0.161 \\
& MambaXray-VL~\cite{Wang_2025_CVPR} & 2025 & \underline{0.422} & 0.268 & 0.184 & 0.133 & 0.289 & 0.167 \\
& MPO~\cite{MPO} & 2025 & 0.416 & 0.269 & 0.191 & 0.139 & 0.309 & 0.162 \\
& PriorRG~\cite{liu2026priorrgpriorguidedcontrastivepretraining} & 2026 & 0.412 & \underline{0.290} & \underline{0.220} & \underline{0.175} & \underline{0.324} & \underline{0.189}  \\
\midrule
& CPVB-RG & ours & \textbf{0.428} & \textbf{0.295} & \textbf{0.223} & \textbf{0.176} & \textbf{0.328} & \textbf{0.196} \\
\midrule
\multirow{16}{*}{\parbox[c]{1.2cm}{\centering IU X-Ray}}
& R2Gen~\cite{r2gen} & 2020 & 0.470 & 0.304 & 0.219 & 0.165 & 0.371 & 0.187 \\
& R2GenCMN~\cite{r2gencmn} & 2021 & 0.475 & 0.309 & 0.222 & 0.170 & 0.375 & 0.191 \\
& PPKED~\cite{ppked} & 2021 & 0.483 & 0.315 & 0.224 & 0.168 & 0.376 & 0.187 \\
& AlignTrans~\cite{aligntrans} & 2021 & 0.484 & 0.313 & 0.225 & 0.173 & 0.379 & 0.204 \\
& CMCL~\cite{cmcl} & 2021 & 0.473 & 0.305 & 0.217 & 0.162 & 0.378 & 0.186 \\
& Clinical-BERT~\cite{clinicalbert} & 2022 & 0.495 & 0.330 & 0.231 & 0.170 & 0.376 & 0.209 \\
& METformer~\cite{metformer} & 2023 & 0.483 & 0.322 & 0.228 & 0.172 & 0.380 & 0.192 \\
& DCL~\cite{dcl} & 2023 & -- & -- & -- & 0.163 & 0.383 & 0.193 \\
& R2GenGPT~\cite{r2gengpt} & 2023 & 0.465 & 0.299 & 0.214 & 0.161 & 0.376 & 0.219 \\
& BootstrappingLLM~\cite{bootllm} & 2024 & 0.499 & 0.323 & 0.238 & 0.184 & 0.390 & 0.208 \\
& MAN~\cite{man} & 2024 & 0.501 & 0.328 & 0.230 & 0.170 & 0.386 & 0.213 \\
& DCG~\cite{DCG} & 2024 & 0.514 & 0.330 & 0.241 & 0.186 & 0.401 & 0.211 \\
& Med-LMM~\cite{medlmm} & 2024 & -- & -- & -- & 0.168 & 0.381 &  0.209 \\
& MambaXray-VL~\cite{Wang_2025_CVPR} & 2025 & 0.491 & 0.330 & 0.241 & 0.185 & 0.371 & 0.216 \\
& MPO~\cite{MPO} & 2025 & \underline{0.548} & \underline{0.383} & \underline{0.278} & \underline{0.209} & \underline{0.415} & \underline{0.224} \\
\midrule
& CPVB-RG & ours & \textbf{0.552} & \textbf{0.394} & \textbf{0.281} & \textbf{0.211} & \textbf{0.423} & \textbf{0.227} \\
\bottomrule
\end{tabular}
\caption{Natural language generation (NLG) results on MIMIC-CXR and IU X-Ray. MambaXray-VL refers to the model introduced in CXPMRG-Bench~\cite{Wang_2025_CVPR}. The best results are shown in \textbf{bold} and the second-best results are \underline{underlined}. ``--'' indicates that the corresponding metric was not reported in the original publication.}
\label{tab:cxr_nlg}
\end{table*}

\subsection{Prior-Aware Radiology Report Generation}
\label{sec:method:JDRG}

Stage 2 uses the aligned encoders and CPVB to condition report generation. In addition to the CPVB representation $\mathbf{F}_b$, multi-layer features from the current-image encoder are aggregated into joint visual tokens $\mathbf{F}_j$. CPVB supplies the latent queries of a Perceiver module, which retrieves complementary evidence from $\mathbf{F}_j$:

{\small
\begin{equation}
\mathbf{F}_a
=
\operatorname{Perceiver}
\left(
\mathbf{F}_j;\mathbf{F}_b
\right).
\end{equation}
}

The CPVB and Perceiver representations are concatenated to form the decoder conditioning sequence:

{\small
\begin{equation}
\mathbf{Z}
=
[\mathbf{F}_b;\mathbf{F}_a].
\end{equation}
}

We use DistilGPT2 as the language decoder. Cross-attention layers allow it to attend to $\mathbf{Z}$ while generating the report. For a report sequence
$\mathbf{r}=\{r_1,r_2,\ldots,r_T\}$, the conditional probability is:

{\small
\begin{equation}
P(\mathbf{r}\mid\mathbf{Z})
=
\prod_{t=1}^{T}
P(r_t\mid r_{<t},\mathbf{Z}).
\end{equation}
}

The generation branch is trained using token-level cross-entropy:

{\small
\begin{equation}
\mathcal{L}_{\mathrm{CE}}
=
-\sum_{t=1}^{T}
\log P
\left(
r_t^{*}\mid r_{<t}^{*},\mathbf{Z}
\right),
\end{equation}
}

where $r_t^{*}$ is the ground-truth token at step $t$. At inference, the decoder receives only the encoded examinations and available clinical context.

\section{Experiments}
\label{sec:experiments}

\subsection{Experimental Setup}

We evaluate CPVB-RG on two public chest X-ray report-generation benchmarks, MIMIC-CXR~\cite{mimiccxr} and IU X-Ray~\cite{IUxray}. Both datasets contain paired chest radiographs and radiology reports, but differ substantially in scale. Further dataset details are provided in the supplementary material.

The common feature dimension is set to \(d=768\). We initialise the vision encoder with RAD-DINO~\cite{perez-garcia_exploring_2025} and the text encoder with CXR-BERT-General~\cite{10.1007/978-3-031-20059-5_1}. DistilGPT2~\cite{sanh2020distilbertdistilledversionbert} is used as the primary language decoder, with alternative decoders examined in the ablation study. CPVB uses 128 anchor tokens, and the maximum report length is 100 tokens. Training follows the two-stage procedure described in Sec.~\ref{sec:method}. Stage 1 is trained for up to 50 epochs with a batch size of 32, while Stage 2 is trained for up to 50 epochs with a batch size of 16. Both stages use AdamW with a learning rate of \(5\times10^{-5}\), together with learning-rate scheduling and early stopping. All experiments are implemented in PyTorch and conducted on a single NVIDIA A100 GPU.

We use BLEU, ROUGE-L and METEOR to measure lexical agreement between generated and reference reports. Clinical efficacy (CE) Precision, Recall and F1 are computed over 14 clinical observations extracted by CheXbert~\cite{chexbert}. Positive and uncertain labels are treated as present, while negative and unmentioned labels are treated as absent. We additionally report F1RadGraph~\cite{f1radgraph}, which measures agreement between clinical entities and relations, and VLScore~\cite{vlscore}, which assesses image--report consistency. Expert-judged report correctness is reported for the 135 clinical cases. Full metric definitions and the clinical review protocol are provided in the supplementary material.

\subsection{Quantitative Results}
\label{sec:exp:quantitative}

Tables~\ref{tab:cxr_nlg} and~\ref{tab:cxr_ce} compare CPVB-RG with recent report-generation methods. On MIMIC-CXR, CPVB-RG obtains 0.428, 0.295, 0.223 and 0.176 from BLEU-1 to BLEU-4, together with 0.328 ROUGE-L and 0.196 METEOR. These are the best NLG results among the compared methods. The gains over PriorRG~\cite{liu2026priorrgpriorguidedcontrastivepretraining} are consistent across all six metrics, although the BLEU-4 improvement is small. On IU X-Ray, CPVB-RG reaches 0.552, 0.394, 0.281 and 0.211 from BLEU-1 to BLEU-4, with 0.423 ROUGE-L and 0.227 METEOR. Across the two datasets, CPVB-RG performs consistently with longitudinal priors on MIMIC-CXR and under the missing-prior setting on IU X-Ray.

\begin{table}[t]
\centering
\small

\begin{tabular}{l|l|ccc}
\toprule
\textbf{Methods} & \textbf{Year} & \textbf{P} & \textbf{R} & \textbf{F1} \\
\midrule
R2Gen~\cite{r2gen} & 2020 & 0.333 & 0.273 & 0.276 \\
R2GenCMN~\cite{r2gencmn} & 2021 & 0.334 & 0.275 & 0.278 \\
Clinical-BERT~\cite{clinicalbert} & 2022 & 0.397 & 0.435 & 0.415 \\
METformer~\cite{metformer} & 2023 & 0.364 & 0.309 & 0.311 \\
DCL~\cite{dcl} & 2023 & 0.471 & 0.352 & 0.373 \\
KiUT~\cite{kiut} & 2023 & 0.371 & 0.318 & 0.321 \\
BootstrappingLLM~\cite{bootllm} & 2024 & 0.465 & \underline{0.482} & 0.473 \\
CoFE~\cite{cofe} & 2024 & 0.489 & 0.370 & 0.405 \\
MAN~\cite{man} & 2024 & 0.411 & 0.398 & 0.389 \\
DCG~\cite{DCG} & 2024 & 0.441 & 0.414 & 0.404 \\
Med-LMM~\cite{medlmm} & 2024 & 0.412 & 0.373 & 0.395 \\
MambaXray-VL~\cite{Wang_2025_CVPR} & 2025 & 0.371 & 0.321 & 0.340 \\
MPO~\cite{MPO} & 2025 & 0.436 & 0.376 & 0.353 \\
PriorRG~\cite{liu2026priorrgpriorguidedcontrastivepretraining} & 2026 & \underline{0.541} & \textbf{0.485} & \underline{0.511} \\
\midrule
CPVB-RG & ours & \textbf{0.552} & \underline{0.482} & \textbf{0.515} \\
\bottomrule
\end{tabular}
\caption{Clinical efficacy (CE) on MIMIC-CXR. P, R and F1 denote Precision, Recall and F1 score, respectively. The best results are shown in bold and the second-best results are underlined.}
\label{tab:cxr_ce}
\end{table}

The CE results in Tab.~\ref{tab:cxr_ce} provide a complementary view of clinical observation agreement. CPVB-RG achieves the highest Precision of 0.552 and F1 of 0.515, with a Recall of 0.482. Relative to PriorRG, Precision increases by 0.011 and F1 by 0.004, while Recall decreases by 0.003. The higher F1 therefore reflects improved agreement across the extracted clinical observations rather than uniformly better performance on every CE component.

We evaluate clinical entity--relation consistency and image--report alignment under the ReXrank~\cite{rexrank} MIMIC-CXR Findings protocol. All methods use the same test cases, reference Findings and metric implementations, with details provided in the supplementary material. As shown in Tab.~\ref{tab:recent_clinical_metrics}, CPVB-RG obtains 0.2847 F1RadGraph and 0.6692 VLScore. It ranks second in F1RadGraph and is 0.0055 below the best VLScore achieved by the 8B UniRG-CXR. Using an 82M-parameter language decoder, CPVB-RG exceeds the 7B MAIRA-2 by 0.0200 in F1RadGraph and 0.0184 in VLScore. PriorRG is the closest decoder-matched prior-aware baseline because both methods use prior examinations and an 82M-parameter decoder. The two methods differ mainly in how they model current--prior interactions. PriorRG uses attention-based temporal fusion, whereas CPVB/PGFA models current--prior interactions through separate progression-difference and shared-spatial branches. Based on 10,000 patient-level paired bootstrap resamples and two-sided bootstrap \(z\)-tests, CPVB-RG improves over PriorRG by 0.0090 in F1RadGraph (95\% CI: [0.0048, 0.0135], \(p<0.0001\)) and 0.0582 in VLScore (95\% CI: [0.0492, 0.0673], \(p<0.0001\)). Both improvements are statistically significant. The supplementary material also reports zero-shot transfer from MIMIC-CXR to IU X-Ray.

\begin{table}[h]
\centering
\footnotesize
\setlength{\tabcolsep}{7pt}
\begin{tabular}{lccc}
\toprule
\textbf{Method} & \textbf{Params} $\downarrow$ & \textbf{F1RadGraph} $\uparrow$ & \textbf{VLScore} $\uparrow$ \\
\midrule
PriorRG~\cite{liu2026priorrgpriorguidedcontrastivepretraining}
    & \textbf{82M} & 0.2757 & 0.6110 \\
CheXagent~\cite{chexagent}
    & 2.7B & 0.2530 & \underline{0.6724} \\
UniRG-CXR~\cite{unirg}
    & 8B & \textbf{0.3040} & \textbf{0.6747} \\
MAIRA-2~\cite{maira2}
    & 7B & 0.2647 & 0.6508 \\
\midrule
CPVB-RG
    & \textbf{82M} & \underline{0.2847} & 0.6692 \\
\bottomrule
\end{tabular}
\caption{Comparison with recent systems using F1RadGraph and VLScore. Params denotes the parameter count of the language backbone or decoder.}
\label{tab:recent_clinical_metrics}
\end{table}

\begin{table*}[t]
\centering
\small
{
\begin{tabular}{lcccc}
\toprule
\textbf{Variant} &
\textbf{BLEU-4} $\uparrow$ &
\textbf{CE F1} $\uparrow$ &
\textbf{F1RadGraph} $\uparrow$ &
\textbf{VLScore} $\uparrow$ \\
\midrule
PriorRG-Style Fusion        & 0.1753 & 0.4892 & 0.2762 & 0.6578 \\
Single Cross-Attention   & 0.1240 & 0.5128 & 0.2224 & 0.6448 \\
w/o PGFA                 & 0.1689 & 0.4739 & 0.2785 & 0.6603 \\
w/o Difference Branch    & 0.1741 & 0.5122 & 0.2839 & 0.6673 \\
w/o Spatial Branch       & 0.1745 & 0.5064 & 0.2767 & 0.6644 \\
Spatial Addition         & 0.1729 & 0.5040 & 0.2783 & 0.6621 \\
\midrule
Full CPVB-RG             & \textbf{0.1760} & \textbf{0.5150} &
                           \textbf{0.2847} & \textbf{0.6692} \\
\bottomrule
\end{tabular}}
\caption{Fine-grained ablation of CPVB and PGFA. BLEU-4 and CE F1 use the standard MIMIC-CXR test set, while F1RadGraph and VLScore follow the ReXrank Findings protocol. PriorRG-Style Fusion adopts PriorRG's temporal attention, with all other settings fixed across variants. Best results are bolded.}
\label{tab:cpvb_fine_ablation}
\end{table*}

\subsection{Ablation Study}

\begin{figure}[h]
  \centering
  \includegraphics[width=0.85\linewidth]{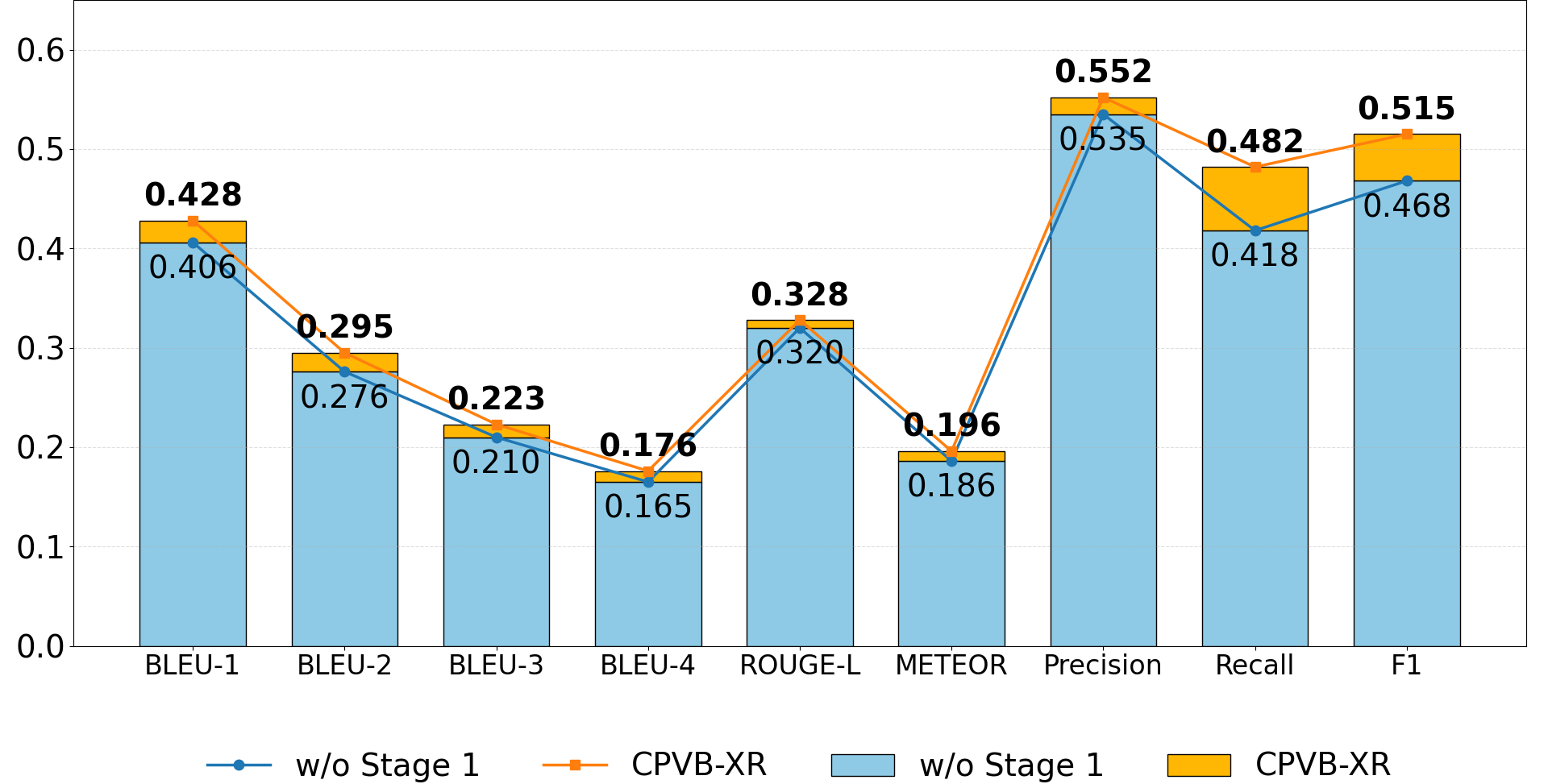}
  \caption{Effect of Stage 1 pre-training on report generation. ``w/o'' denotes ``without''.}
  \label{st1}
\end{figure}

\subsubsection{Effectiveness of Stage 1 Pre-training}

Fig.~\ref{st1} compares the complete model with a variant trained without Stage 1. Removing prior-guided image-text alignment reduces every reported NLG and CE metric. The larger decreases in Recall and F1 indicate that Stage 1 mainly helps the generation model retain clinical observations, while the accompanying NLG reductions show that its effect is not limited to the CE evaluation.

\begin{figure*}[ht]
  \centering
  \includegraphics[width=0.8\linewidth]{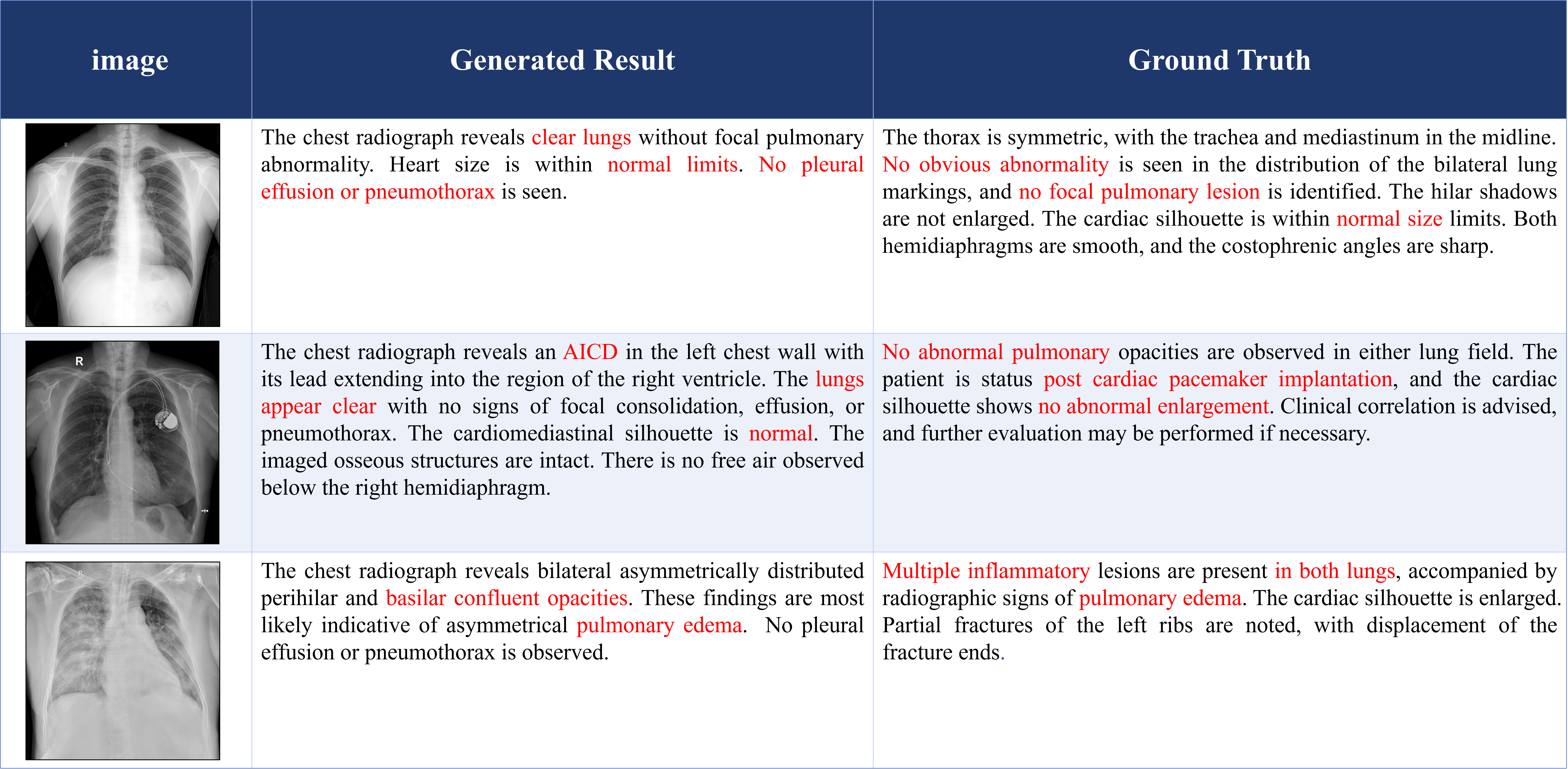}
  \caption{
Examples of generated radiology reports and their reference reports for clinical chest X-ray cases. Text in \textcolor{red}{red} marks clinically important content in the generated and reference reports for comparison.
}
\label{fig:real_sample_2}
\end{figure*}

\subsubsection{Ablation on Input Configuration}

Tab.~\ref{tab:input_ablation} examines the effect of progressively introducing prior examinations, temporal-order information and clinical context. Using only the current examination gives 0.152 BLEU-4 and 0.311 CE F1. Adding an unordered prior does not improve performance, while encoding temporal order raises the scores to 0.158 and 0.347. Clinical context further improves them to 0.168 and 0.471. Applying CPVB to the complete input configuration achieves the best results of 0.176 BLEU-4 and 0.515 CE F1.

\subsubsection{Fine-Grained Ablation of CPVB and PGFA}
\label{sec:exp:cpvb_ablation}

Having established the contribution of CPVB, we examine its internal design under controlled settings. We keep the input data, visual encoder, language decoder, training objectives and optimisation settings fixed, varying the current--prior fusion mechanism. PriorRG-Style Fusion replaces PGFA with the attention-based temporal fusion used in PriorRG, while Single Cross-Attention retains only one of the two Cross-Attention units. The remaining variants remove PGFA, its progression-difference branch or its spatial-guidance branch. Spatial Addition replaces attention-map subtraction with addition. As shown in Tab.~\ref{tab:cpvb_fine_ablation}, full CPVB-RG achieves the best results across all four metrics. Removing PGFA reduces BLEU-4, CE F1, F1RadGraph and VLScore by 0.0071, 0.0411, 0.0062 and 0.0089, respectively. Removing the Difference Branch causes smaller but consistent decreases, while removing the Spatial Branch lowers F1RadGraph and VLScore by 0.0080 and 0.0048.

\begin{table}[t]
\centering
\small
\setlength{\tabcolsep}{4pt}
\begin{tabular}{c c c c c | c c}
\toprule
\textbf{Current} & \textbf{Prior} & \textbf{Order} & \textbf{Text} & \textbf{CPVB} & \textbf{BLEU-4} & \textbf{CE F1} \\
\midrule
\checkmark & --         & --         & --         & --         & 0.152 & 0.311 \\
\checkmark & \checkmark & --         & --         & --         & 0.149 & 0.298 \\
\checkmark & \checkmark & \checkmark & --         & --         & 0.158 & 0.347 \\
\checkmark & \checkmark & \checkmark & \checkmark & --         & 0.168 & 0.471 \\
\checkmark & \checkmark & \checkmark & \checkmark & \checkmark & \textbf{0.176} & \textbf{0.515} \\
\bottomrule
\end{tabular}
\caption{Ablation of the input configuration and CPVB. Order denotes temporal-order information, and Text denotes the available clinical context. The best results are shown in bold.}
\label{tab:input_ablation}
\end{table}

Replacing attention-map subtraction with addition reduces all four metrics, including decreases of 0.0110 in CE F1, 0.0064 in F1RadGraph and 0.0071 in VLScore. This result supports attention-map subtraction as a design choice and is consistent with its intended role of reducing shared temporal responses and emphasising interval-change evidence. PriorRG-Style Fusion obtains a similar BLEU-4 score, but its CE F1, F1RadGraph and VLScore are lower by 0.0258, 0.0085 and 0.0114, respectively, showing that the benefit of CPVB/PGFA extends beyond lexical overlap. Retaining only one Cross-Attention unit substantially reduces BLEU-4, F1RadGraph and VLScore. Together, these results support the progression-difference branch, spatial guidance, attention-map subtraction and both Cross-Attention units. Additional experiments in the supplementary material examine decoder selection and the reduction factor \(r\). DistilGPT2 provides the best observed balance between generation performance and decoder size, while \(r=2\) gives the strongest overall CPVB result.

\subsection{Clinical Case Analysis}
\label{sec:exp:cca}

To examine performance on real clinical data, three collaborating radiologists evaluated 135 de-identified chest X-ray cases selected through stratified random sampling across 14 clinical observation categories. The sample was designed to cover diverse clinical findings rather than reflect their natural prevalence. One radiologist performed the initial assessment of each generated report, and two others reviewed it. This non-interventional secondary-data study was approved by the ethics committees of the authors' institution and the source clinical institution. The research team had no access to direct patient identifiers or re-identification keys. Further details on reader experience, case selection, review procedure and assessment criteria are provided in the supplementary material. As shown in Tab.~\ref{tab:realworld_results}, 89.6\% of the reports in this stratified evaluation set were judged overall correct. This includes 60.7\% that were completely correct and 28.9\% that contained only minor omissions. A further 5.93\% contained correct findings, incorrect statements and/or important omissions, while 4.44\% were completely incorrect. These results are preliminary given the sample size, but they permit analysis of the model's successes and remaining errors on clinical cases.

\begin{table}[t]
\centering
\small
\begin{tabular}{lc}
\hline
\textbf{Category} & \textbf{Proportion (\%)} \\
\hline
Overall correct & 89.6 \\
\hline
Completely correct reports & 60.7 \\
Partially correct & 28.9 \\
Mixed \& Missing & 5.93 \\
Completely incorrect & 4.44 \\
\hline
\end{tabular}
\caption{
Report-level evaluation of 135 chest X-ray cases. Overall correct is the sum of completely and partially correct reports. Partially correct denotes reports with correct information and minor omissions. Mixed \& Missing denotes reports containing correct and incorrect statements and/or missing important findings. Completely incorrect denotes no correct information.
}
\label{tab:realworld_results}
\end{table}

Fig.~\ref{fig:real_sample_2} shows three representative cases. The first agrees with the normal reference, reporting clear lungs, normal heart size and no pleural effusion or pneumothorax. The second identifies an implanted cardiac device and no pulmonary abnormality, but calls the device an AICD rather than a pacemaker and adds details absent from the reference. The third captures pulmonary oedema and opacities but describes basilar confluent opacities instead of the reported multiple bilateral inflammatory opacities. These cases reveal errors in device classification, disease distribution and diagnostic specificity.

\section{Conclusion}

We presented CPVB-RG, a prior-aware framework designed to distinguish progression-related differences from shared spatial responses and connect this longitudinal evidence with clinical context through prior-guided image-text alignment and lightweight autoregressive decoding. Experiments on MIMIC-CXR~\cite{mimiccxr} and IU X-Ray~\cite{IUxray} show improvements across natural language generation and clinically grounded metrics. Controlled ablations support the design, while evaluation on 135 clinical cases reveals remaining errors in device classification, disease distribution and diagnostic specificity. Future work will examine irregular or missing priors and multi-institutional validation.

{
    \small
    \bibliographystyle{ieeenat_fullname}
    \bibliography{main}
}

\end{document}